\pdfoutput=1
\documentclass{article}

\usepackage{microtype}
\usepackage{graphicx}
\usepackage{subfigure}
\usepackage{booktabs} 
\usepackage[draft, cachedir=.]{minted} 
\usepackage{listings}
\usepackage{enumitem}

\usepackage{hyperref}

\usepackage[accepted]{whi2020}

\icmltitlerunning{Vizarel: A System to Help Better Understand RL Agents}

\begin{document}

\twocolumn[
    \icmltitle{Vizarel: A System to Help Better Understand RL Agents}



\icmlsetsymbol{equal}{*}

\begin{icmlauthorlist}
\icmlauthor{Shuby Deshpande}{cmu}
\icmlauthor{Jeff Schneider}{cmu}
\end{icmlauthorlist}

\icmlaffiliation{cmu}{Carnegie Mellon University, Pittsburgh, PA, United States}

\icmlcorrespondingauthor{Shuby Deshpande}{shubhand@cs.cmu.edu}

\icmlkeywords{Reinforcement Learning, Interpretability, Visualization}

\vskip 0.3in
]



\printAffiliationsAndNotice{\icmlEqualContribution} 

\begin{abstract}
Visualization tools for \emph{supervised learning} have allowed users to interpret, introspect, and gain intuition for the successes and failures of their models. While \emph{reinforcement learning} practitioners ask many of the same questions, existing tools are not applicable to the RL setting. In this work, we describe our initial attempt at constructing a prototype of these ideas, through identifying possible features that such a system should encapsulate. Our design is motivated by envisioning the system to be a platform on which to experiment with interpretable reinforcement learning.
\end{abstract}


\section{Introduction}
\label{intro}

Machine learning systems have made impressive advances due to their ability to learn high dimensional models from large amounts of data \cite{LeCun_Bengio_Hinton_2015}. However,  high dimensional models are hard to understand and trust \cite{doshivelez2017rigorous}. Visualization systems are important for overcoming these challenges.

Many tools exist for addressing these challenges in the \emph{supervised learning} setting.  which find usage in tracking metrics \cite{satyanarayan_vegalite}, generating graphs of model internals \cite{wongsuphasawat_tfgraphs}, and visualizing embeddings \cite{maaten_tsne}. However, there is no corresponding set of tools for the reinforcement learning setting. At first glance, we may repurpose existing libraries or packages for this task. However, we quickly run into limitations, which arise due to the intent with which tools were designed in the first place.

Reinforcement learning is fundamentally an interactive science \cite{Neftci_Averbeck_2019} in that there is a stronger feedback loop between the researcher and model (i.e. agent), compared to supervised learning. We need tools that reflect this dynamic instead of limiting us to the constraints imposed by the supervised learning framework.


Visualization systems at their core consist of two components: representation and interaction. Though these may appear to be disparate, it is hard to discount the influence that each has on each other. The tools we use for representation affect how we interact with the system, and our interaction affects the representations that we create \cite{Yi_understanding_interaction}. Visualization interfaces should adhere to the human action cycle \cite{norman_2013}, which provides us with a useful model to think about when designing features which our systems should encapsulate.


\nocite{Bostock_Ogievetsky_Heer_2011}
\nocite{Pike_Stasko_2009}

Three dimensions along which to evaluate interaction in visualization systems, as proposed by \cite{Beaudouin-Lafon_2004}, and adapted here for relevance, are:
\begin{itemize}[noitemsep, nolistsep]
    \item[-] \textit{descriptive power}: the ability to describe a significant range of existing interfaces
    \item[-] \textit{evaluative power}: the ability to help assess multiple alternatives
    \item[-] \textit{generative power}: the ability to help create new designs
\end{itemize}

\nocite{tensorboard}

Existing tools primarily focus on discriptive power. Using them, we can plot common descriptive metrics such as cumulative reward, TD-error, action values, to name a few. However, these systems either lack or are deficient in evaluative and generative power. Ideally, the systems we use should help us answer questions such as:
\begin{itemize}[noitemsep, nolistsep]
    \item[-] What sequence of dynamics causes my agent to behave the way it does?
    \item[-] What actions should I take to induce the intended agent behavior instead?
    \item[-] What effects does experiencing noteworthy states have on the resulting policy? Are there other states which lead to similar outcomes?
\end{itemize}
These are far from an exhaustive list of questions that the researcher may pose during training agent policies but are chosen to illustrate the current gap that our interfaces face with regards to evaluative and generative power.
This paper describes our initial attempt at constructing a visualization system that can answer these questions. Concretely,  we make the following contributions:
\begin{itemize}[noitemsep, nolistsep]
    \item identifying features in which an interactive system for interpretable reinforcement learning should~encapsulate.
    \item building a prototype of these ideas, which instantiates a working system with these features
    \item enumerating upcoming features which extend the system's existing capabilities
\end{itemize}

\begin{figure}[hbtp!]
    \centering
    \includegraphics[width=\linewidth]{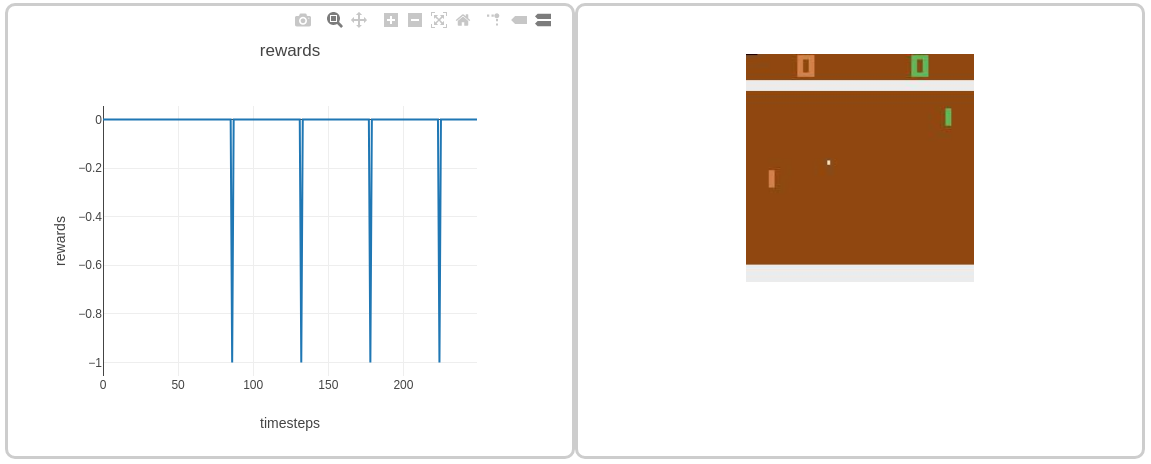}
    \caption{\textbf{Reward + State Space Viewport} 
    Visualizing autogenerated reward \& state space viewports for the Pong environment. This representation should provide the user better intuition about the correspondence between rewards and states, especially for environments with denser rewards.
    \label{fig:reward_plot}}
    \vspace{-1em}
\end{figure}

\section{Preliminaries} \label{formulation}
We use the standard reinforcement learning setup~\cite{sutton}. An agent interacting with an environment at discrete timesteps $t$, receiving a scalar reward~$r_t$.
The agent's behavior is defined by a policy $\pi$, which maps states $s \in S$, to a probability distribution over actions, $\pi \rightarrow P(A)$. The environment $E$ can be stochastic, which is modeled by a Markov decision process with a state space $S$, action space $A \in R^n$, an initial state distribution $p(s_0)$, a transition function $p(s_{t+1} \mid s_{t}, a)$, and a reward function $r(s_t, a_t)$. The return from a state is defined as $R_t = \sum_{i=t}^{T} \gamma^{i-t}r(s_t,a_t)$, with a discount factor $\gamma \in [0,1]$.
We use a replay buffer \cite{mnih2013playing} to store the agent's experiences $e_t = (s_t, a_t, r_t, s_{t+1})$ in a buffer $B = \{e_0, e_1, ..., e_T\}$.

\section{System Description}

Our tool has two components, the frontend dashboard and control panel, and the backend storage server and logging~unit.

\subsection{Frontend}
The frontend enables the construction of multiple viewports, which serve as the base class for further visualization extensions. Each viewport as an abstract entity can be backed by different specs, based on the underlying data stream. For example, one could use:
\begin{enumerate}[noitemsep, nolistsep]
    \item \textit{image buffers}: to visualize image based observation spaces (or non-image based spaces if rendering is enabled)
    \item \textit{line plots}: to visualize non-image based spaces, action values, and rewards
    \item \textit{scatter plots}: to visualize embedding spaces
\end{enumerate}
This naturally leads to the idea of an ecosystem of plugins that can be integrated into the core system to support different visualization schemes and algorithms. Though these are the currently available viewports, in a later section we describe upcoming viewport designs that are being integrated, to support additional visualizations.
The following subsections detail different views that the frontend interface currently supports.
\subsubsection{State Spaces}
Referring to the state-space formulation from \S \ref{formulation}, states can primarily be classified as either image-based or non-image based spaces. The type of observation space influences the corresponding spec through which the viewport is generated. We provide two examples that illustrate how these differing specs can result in different viewports. Consider a non-image based observation space, such as that for the inverted pendulum task. Here, the state vector $\vec{s} = \{\sin(\theta), \cos(\theta), \dot{\theta}\}$, where $\theta$ is the angle which the pendulum makes with the vertical.

\nocite{openai_gym}

\begin{figure}[t]
    \centering
    \includegraphics[width=\linewidth]{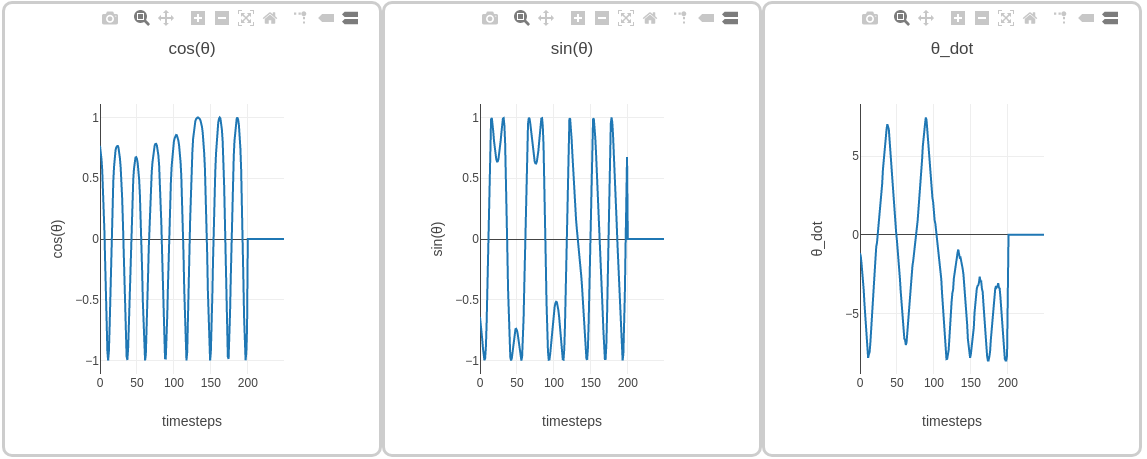}
    \vspace{-0.5em}
    \caption{\textbf{State Space Viewport}:
    \small
    Visualizing autogenerated state space viewports for the inverted pendulum task. This representation with an rendered image overlay (see Figure \ref{fig:action_space}), provides the user with better intuition about the correspondence between state dimensions and images, which humans find easier to interpret.} 
    \label{fig:state_space_viewport}
    \vspace{-1em}
\end{figure}

We can visualize the state vector components individually, which gives us a sense of how states vary across episode timesteps (Figure \ref{fig:state_space_viewport}). Since an image representation is easier for humans to interpret, it seems reasonable to also generate an additional viewport which tracks the corresponding changes in image space. Having this simultaneous visualization is useful since this now enables us to jump back and forth between the state representation which the agent receives, and the corresponding element in image space, by simply clicking on the desired timestep in the state viewport.

\begin{figure}[t]
    \centering
    \includegraphics[width=\linewidth]{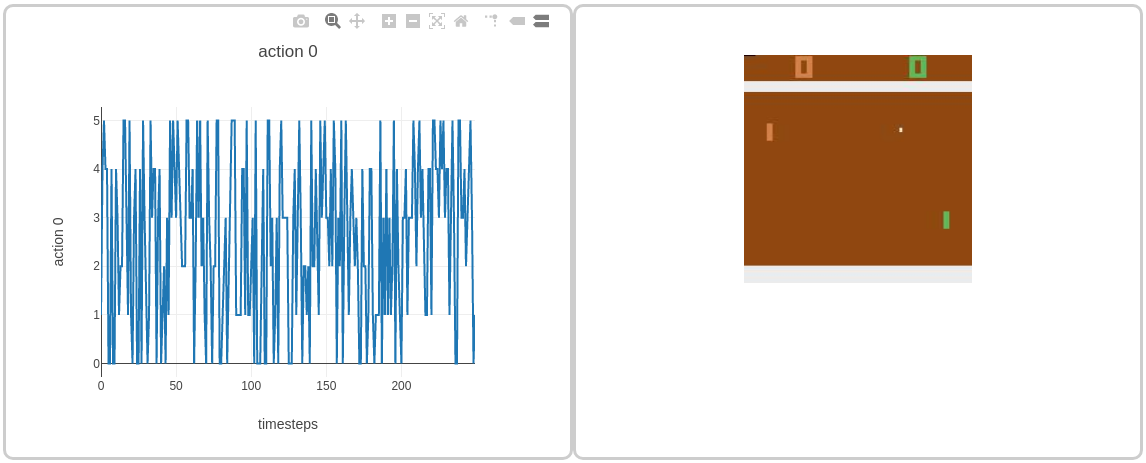}
    \caption{\textbf{Action Spaces}:
    \label{fig:action_space}
    Visualizing autogenerated action space viewports for the Pong environment. This representation with an image overlay, provides the user better intuition about the current agent policy. This along with a slider to control and query episode level logs (see \S 3.1.5), can help the user to better debug agent~policies.}
    \vspace{-1em}
\end{figure}

For environments that have higher dimensional non image states, such as that of a robotic arm with multiple degrees of freedom, we could visualize individual state components. However, since this may not be intuitive, we can also generate an additional viewport as an overlay to display an image rendering of the environment, similar to that shown in Figure \ref{fig:action_space}.

\subsubsection{Action Spaces}
As per the action space formulation from \S \ref{formulation}, at each timestep $t$ the agent chooses an action $a_t$, which is either discrete or continuous depending on the type of action space. We can visualize how the action $a_t$ varies across the episode by creating a viewport backed by a line plot spec. For agents where we have access to a distribution over actions instead, we can generate a viewport backed by a histogram spec, and visualize how the action distribution changes over time. A similar visualization can be generated for agents that make use of action-value functions \cite{sutton}, for action selection.

\subsubsection{Rewards}
As per the reward formulation from \S \ref{formulation}, at each timestep $t$, the agent receives a reward $r_t$ conditioned on the previous state $s_{t-1}$ and action $a_{t-1}$. The reward is typically a scalar quantity, so it would be useful to generate a viewport backed by a line plot spec.

For most agent environments, the reward function comprises of different components weighted by different coefficients. These individual components are often easier to interpret since they are usually backed by a physically motivated quantity tied to specific behaviors that we wish to either reward or penalize. In situations where we have access to these, we can autogenerate multiple viewports each of which visualizes different components of this reward function~vector.

\subsubsection{Replay Buffer}
As formulated in \S \ref{formulation}, the replay buffer stores the agent's experiences $e_t = (s_t, a_t, r_t, s_{t+1})$ in a buffer $B = \{e_0, e_1, ..., e_T\} \forall i \in [0, T]$.
For off-policy algorithms, the replay buffer is of crucial importance, since it in effect serves as a proxy dataset during agent policy updates. For visualizing datasets, there exist tools, which provide the user with an intuition for the underlying data distribution. Similarly, it would make sense to visualize the replay buffer state, since this in effect a proxy to a dataset for the RL agent.
\nocite{Facets}

Since the individual elements of the replay buffer are at least a four-dimensional vector, this rules out the possibility of generating viewports backed by specs, in the original space. We can instead generate a lower-dimensional projection of the replay buffer distribution, which provides a notion of the replay buffer diversity.

\begin{figure}[t]
    \centering
    \includegraphics[width=\linewidth]{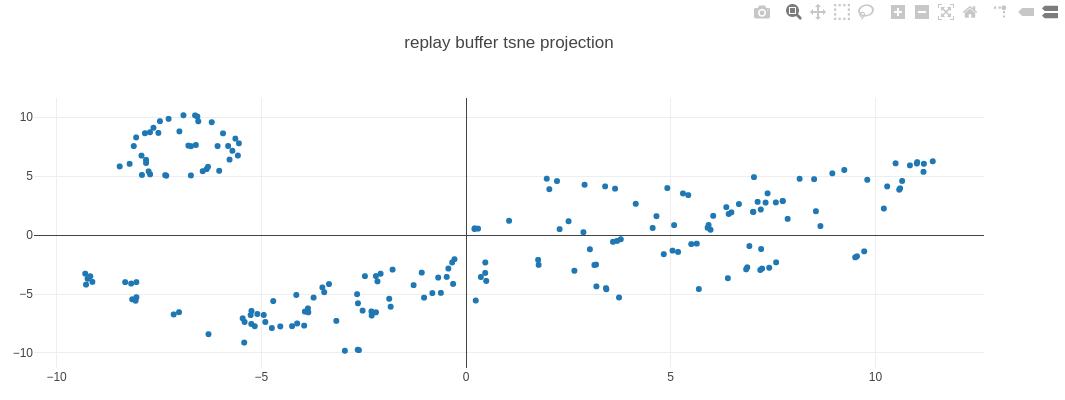}
    \caption{\textbf{Replay Buffer Projection}
    Projecting the contents of the replay buffer into a 2D space for easier navigation. This provides a proxy to the replay buffer diversity, and can help in subsequent debugging.
    \label{fig:rb_projection}}
    \vspace{-1em}
\end{figure}

This is supported by the current system, which computes a lower-dimensional projection \cite{maaten_tsne}, of the replay buffer, and then allows the user to visualize the distribution, along with a hover icon which describes the original 4-tuple from which the projected point was computed.

\subsubsection{Control Panel}

\begin{figure}[hbtp!]
    \centering
    \includegraphics[width=\linewidth]{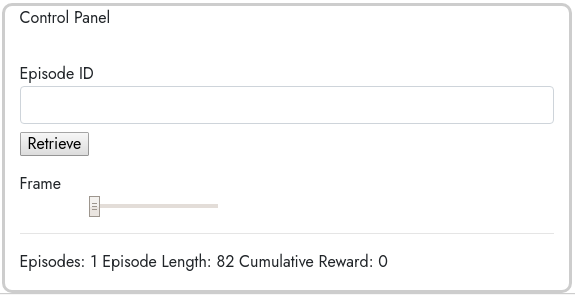}
    \caption{Control Panel}
    \label{fig:control_panel}
\end{figure}

The control panel (Figure \ref{fig:control_panel}) is a common component across all views which supports functionality to:
\begin{enumerate}[noitemsep, nolistsep]
    \item display high-level descriptive metrics such as average return, average time per episode, and the number of episodes.
    \item retrieve logs for arbitrary episode IDs.
    \item control the currently active frame, which is reflected in the corresponding state, action, and reward viewports.
\end{enumerate}
Possible extensions to this are to provide real-time suggestions to the user, to help navigation through a large collection of episode logs.

\subsection{Backend}
The backend system is responsible for storage, logging, and communication with one or multiple frontend clients attempting to interface with the agent.

\begin{figure}[t]
    \centering
    \includegraphics[width=\linewidth]{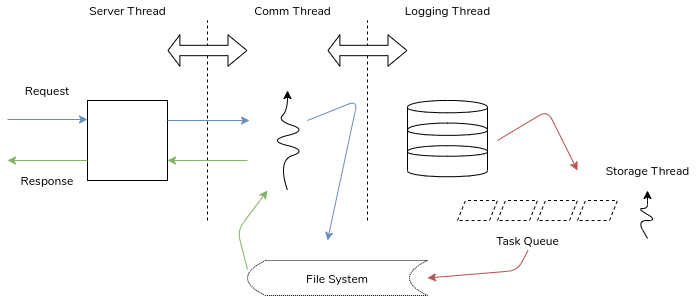}
    \caption{Backend Architecture Overview}
    \label{fig:my_label}
\end{figure}

At a high level, it consists of three sub-components: the serving thread, the communication thread, and the logging thread. The serving thread interfaces with frontend clients, which request data streams for visualization. The communication thread acts as the arbiter between the serving and logging threads, performs the logical mapping from the server request to the data store, and communicates with the logging thread to notify it of validated commands received from the frontend.

The logging thread is responsible for caching tensors to the data store. It does so by pushing data onto a task queue, which is then asynchronously committed to disk by another thread, after running storage optimizations, designed as such to reduce the computation overhead within the main agent training loop.

The overhead of integrating the overall system is minimal and can be enabled through a mere 2 lines of code, one for initializing the system, and another for caching tensors within the agent training loop as shown in Figure \ref{fig:code}.
\nocite{garage}

\begin{figure}
    \centering
    \begin{minted}[breaklines]{python}
    from vizarel.container import VizarelState
    
    logger = VizarelState(steps, obs_dim, obs_type, action_dim, action_type, reward_dim, reward_type)
    
    logger.log_state(n_samples, obses, actions, rewards, dones)
    \end{minted}
    \caption{Sample code to enable logging}
    \label{fig:code}
\end{figure}

\section{Future Work}

This paper describes the preliminary version of the system we have prototyped as a testbed for ideas. There are multiple features under development that contribute towards both the core interface and the plugin ecosystem which was alluded to earlier. We enumerate some below as a representative sample:
\begin{itemize}[noitemsep, nolistsep]
    \item[-] Dynamically switching logging on or off, conditional on the occurrence of noteworthy experiences during agent training.
    \item[-] Integration of additional data streams such as saliency maps \cite{greydanus2017visualizing} for image-based state spaces, which can be enabled through the plugin ecosystem.
    \item[-] Data processing before rendering, for example, chaining different action dimensions, or clustering rewards across time to diagnose similar states.
\end{itemize}

These are features that we think would be useful to have, but we expect that the best features yet to be built will emerge through feedback from the broader RL and ML interpretability communities.


\section*{Acknowledgements}

We thank Benjamin Eysenbach for valuable discussions and feedback over the initial drafts of this work. This work is supported by the CMU Argo AI Center. Any opinions, recommendations, and conclusions expressed in this material are those of the author(s) and do not reflect the views of any funding agencies.

\bibliography{references}

\begin{thebibliography}{18}
\providecommand{\natexlab}[1]{#1}
\providecommand{\url}[1]{\texttt{#1}}
\expandafter\ifx\csname urlstyle\endcsname\relax
  \providecommand{\doi}[1]{doi: #1}\else
  \providecommand{\doi}{doi: \begingroup \urlstyle{rm}\Url}\fi

\bibitem[gar(2019)]{garage}
Garage: A toolkit for reproducible reinforcement learning research.
\newblock \url{https://github.com/rlworkgroup/garage}, 2019.

\bibitem[Fac(2020)]{Facets}
Facets: Visualizations for machine learning datasets.
\newblock \url={https://pair-code.github.io/facets/}, 2020.

\bibitem[ten(2020)]{tensorboard}
Tensorboard: Tensorflow's visualization toolkit.
\newblock \url{https://github.com/tensorflow/tensorboard}, 2020.

\bibitem[Beaudouin-Lafon(2004)]{Beaudouin-Lafon_2004}
Beaudouin-Lafon, M.
\newblock Designing interaction, not interfaces.
\newblock \emph{Proceedings of the working conference on Advanced visual
  interfaces - AVI ’04}, 2004.
\newblock \doi{10.1145/989863.989865}.

\bibitem[Bostock et~al.(2011)Bostock, Ogievetsky, and
  Heer]{Bostock_Ogievetsky_Heer_2011}
Bostock, M., Ogievetsky, V., and Heer, J.
\newblock D3 data-driven documents.
\newblock \emph{IEEE Transactions on Visualization and Computer Graphics},
  17\penalty0 (12):\penalty0 2301–2309, Dec 2011.
\newblock \doi{10.1109/tvcg.2011.185}.

\bibitem[Brockman et~al.(2016)Brockman, Cheung, Pettersson, Schneider,
  Schulman, Tang, and Zaremba]{openai_gym}
Brockman, G., Cheung, V., Pettersson, L., Schneider, J., Schulman, J., Tang,
  J., and Zaremba, W.
\newblock Openai gym, 2016.

\bibitem[Doshi-Velez \& Kim(2017)Doshi-Velez and Kim]{doshivelez2017rigorous}
Doshi-Velez, F. and Kim, B.
\newblock Towards a rigorous science of interpretable machine learning, 2017.

\bibitem[Greydanus et~al.(2017)Greydanus, Koul, Dodge, and
  Fern]{greydanus2017visualizing}
Greydanus, S., Koul, A., Dodge, J., and Fern, A.
\newblock Visualizing and understanding atari agents, 2017.

\bibitem[LeCun et~al.(2015)LeCun, Bengio, and Hinton]{LeCun_Bengio_Hinton_2015}
LeCun, Y., Bengio, Y., and Hinton, G.
\newblock Deep learning.
\newblock \emph{Nature}, 521\penalty0 (7553):\penalty0 436–444, May 2015.
\newblock \doi{10.1038/nature14539}.
\newblock URL \url{https://www.nature.com/articles/nature14539}.

\bibitem[Maaten \& Hinton(2008)Maaten and Hinton]{maaten_tsne}
Maaten, L. V.~D. and Hinton, G.
\newblock Visualizing data using t-sne.
\newblock \emph{Journal of Machine Learning Research}, 9:\penalty0 2579–2605,
  2008.
\newblock URL
  \url{https://lvdmaaten.github.io/publications/papers/JMLR_2008.pdf}.

\bibitem[Mnih et~al.(2013)Mnih, Kavukcuoglu, Silver, Graves, Antonoglou,
  Wierstra, and Riedmiller]{mnih2013playing}
Mnih, V., Kavukcuoglu, K., Silver, D., Graves, A., Antonoglou, I., Wierstra,
  D., and Riedmiller, M.
\newblock Playing atari with deep reinforcement learning, 2013.

\bibitem[Neftci \& Averbeck(2019)Neftci and Averbeck]{Neftci_Averbeck_2019}
Neftci, E.~O. and Averbeck, B.~B.
\newblock Reinforcement learning in artificial and biological systems.
\newblock \emph{Nature Machine Intelligence}, 1\penalty0 (3):\penalty0
  133–143, Mar 2019.
\newblock \doi{10.1038/s42256-019-0025-4}.
\newblock URL \url{https://www.nature.com/articles/s42256-019-0025-4}.

\bibitem[Norman(2013)]{norman_2013}
Norman, D.~A.
\newblock \emph{The design of everyday things}.
\newblock The MIT Press, 2013.

\bibitem[Pike et~al.(2009)Pike, Stasko, Chang, and
  O’Connell]{Pike_Stasko_2009}
Pike, W.~A., Stasko, J., Chang, R., and O’Connell, T.~A.
\newblock The science of interaction.
\newblock \emph{Information Visualization}, 8\penalty0 (4):\penalty0 263–274,
  Jan 2009.
\newblock \doi{10.1057/ivs.2009.22}.
\newblock URL \url{https://www.cc.gatech.edu/~john.stasko/papers/iv09-soi.pdf}.

\bibitem[Satyanarayan et~al.(2017)Satyanarayan, Moritz, Wongsuphasawat, and
  Heer]{satyanarayan_vegalite}
Satyanarayan, A., Moritz, D., Wongsuphasawat, K., and Heer, J.
\newblock Vega-lite: A grammar of interactive graphics.
\newblock \emph{IEEE Transactions on Visualization and Computer Graphics},
  23\penalty0 (1):\penalty0 341–350, Jan 2017.
\newblock \doi{10.1109/tvcg.2016.2599030}.

\bibitem[Sutton \& Barto(2018)Sutton and Barto]{sutton}
Sutton, R.~S. and Barto, A.~G.
\newblock \emph{Introduction to reinforcement learning}.
\newblock The MIT Press, 2018.

\bibitem[Wongsuphasawat et~al.(2018)Wongsuphasawat, Smilkov, Wexler, Wilson,
  Mane, Fritz, Krishnan, Viegas, and Wattenberg]{wongsuphasawat_tfgraphs}
Wongsuphasawat, K., Smilkov, D., Wexler, J., Wilson, J., Mane, D., Fritz, D.,
  Krishnan, D., Viegas, F.~B., and Wattenberg, M.
\newblock Visualizing dataflow graphs of deep learning models in tensorflow.
\newblock \emph{IEEE Transactions on Visualization and Computer Graphics},
  24\penalty0 (1):\penalty0 1–12, Jan 2018.
\newblock \doi{10.1109/tvcg.2017.2744878}.
\newblock URL
  \url{http://idl.cs.washington.edu/files/2018-TensorFlowGraph-VAST.pdf}.

\bibitem[Yi et~al.(2007)Yi, Kang, Stasko, and
  Jacko]{Yi_understanding_interaction}
Yi, J.~S., Kang, Y.~a., Stasko, J., and Jacko, J.
\newblock Toward a deeper understanding of the role of interaction in
  information visualization.
\newblock \emph{IEEE Transactions on Visualization and Computer Graphics},
  13\penalty0 (6):\penalty0 1224–1231, Nov 2007.
\newblock \doi{10.1109/tvcg.2007.70515}.

\end{thebibliography}
\bibliographystyle{icml2020}

\end{document}